\documentclass[sigconf]{acmart}

\usepackage[ruled,linesnumbered]{algorithm2e}
\usepackage{bm}

\AtBeginDocument{%
  }

\copyrightyear{2025}
\acmYear{2025}
\setcopyright{acmlicensed}
\acmConference[MM '25] {Proceedings of the 33rd ACM International Conference on Multimedia}{October 27--31, 2025}{Dublin, Ireland.}
\acmBooktitle{Proceedings of the 33rd ACM International Conference on Multimedia (MM '25), October 27--31, 2025, Dublin, Ireland}
\acmISBN{979-8-4007-2035-2/2025/10}
\acmDOI{10.1145/3746027.3755478}

\begin{document}

\title{Diffusion-based Adversarial Identity Manipulation for Facial Privacy Protection}

\author{Liqin Wang}
\orcid{0009-0009-2039-9091}
\email{wanglq37@mail2.sysu.edu.cn}
\affiliation{%
\institution{School of Computer Science and \\ Engineering, Sun Yat-sen University} 
\city{Guangzhou}
\country{China}
}

\author{Qianyue Hu}
\orcid{0009-0001-5420-571X}
\email{huqy56@mail3.sysu.edu.cn}
\affiliation{%
\institution{School of Computer Science and \\ Engineering, Sun Yat-sen University} 
\city{Guangzhou}
\country{China}
}

\author{Wei Lu}
\orcid{0000-0002-4068-1766}
\authornote{Corresponding author.}
\email{luwei3@mail.sysu.edu.cn}
\affiliation{
\institution{School of Computer Science and \\ Engineering, Sun Yat-sen University}
\city{Guangzhou}
\country{China}
}

\author{Xiangyang Luo}
\orcid{0000-0003-3225-4649}
\email{luoxy\_ieu@sina.com}
\affiliation{
\institution{State Key Laboratory of Mathematical \\ Engineering and Advanced \\ Computing, Zhengzhou, China}
\city{}
\country{}
} 

\renewcommand{\shortauthors}{Liqin Wang et al.}

\begin{abstract}
The success of face recognition (FR) systems has led to serious privacy concerns due to potential unauthorized surveillance and user tracking on social networks. Existing methods for enhancing privacy fail to generate natural face images that can protect facial privacy. In this paper, we propose diffusion-based adversarial identity manipulation (DiffAIM) to generate natural and highly transferable adversarial faces against malicious FR systems. To be specific, we manipulate facial identity within the low-dimensional latent space of a diffusion model. This involves iteratively injecting gradient-based adversarial identity guidance during the reverse diffusion process, progressively steering the generation toward the desired adversarial faces. The guidance is optimized for identity convergence towards a target while promoting semantic divergence from the source, facilitating effective impersonation while maintaining visual naturalness. We further incorporate structure-preserving regularization to preserve facial structure consistency during manipulation. Extensive experiments on both face verification and identification tasks demonstrate that compared with the state-of-the-art, DiffAIM achieves stronger black-box attack transferability while maintaining superior visual quality. We also demonstrate the effectiveness of the proposed approach for commercial FR APIs, including Face++ and Aliyun.
\end{abstract}







\maketitle


\section{Introduction}
Face recognition (FR) systems~\cite{parkhi2015deep,wang2021deep} have achieved remarkable performance in verification and identification tasks, often surpassing human capabilities across applications like security~\cite{wang2017face}, biometrics~\cite{meden2021privacy}, and forensic investigations~\cite{phillips2018face}. However, the widespread adoption of FR systems has raised serious privacy concerns, particularly due to their potential capacity for mass surveillance~\cite{ahern2007over,wenger2023sok}. For instance, government and private entities can use proprietary FR systems to track user relationships and activities by analyzing face images from social media platforms~\cite{hill2022secretive,shoshitaishvili2015portrait}. Therefore, there is an urgent need for an effective approach to protect facial privacy against black-box FR systems.

Recent works have explored noise-based adversarial examples~\cite{zhou2025numbod,zhou2024darksam} to protect facial privacy by introducing adversarial perturbations to source images to deceive FR systems~\cite{rajabi2021practicality,szegedy2013intriguing,tipim}. However, these adversarial perturbations are typically constrained to $l_p$ norm in the pixel space, resulting in visible artifacts and poor image quality. Unlike noise-based methods, unrestricted adversarial examples~\cite{amtgan,clip2protect,advmakeup,zhu2019generating} are not confined by perturbation budgets, allowing for more natural modifications~\cite{bhattad2019unrestricted,song2018constructing,xiao2018spatially}.

Several efforts have been made to generate unrestricted adversarial examples to mislead malicious FR systems. Among these, makeup-based methods~\cite{amtgan,clip2protect,sun2024diffam} employ generative models to create adversarial faces via makeup transfer. However, these makeup-based methods often require specific makeup references and retraining for new target identities or makeup styles. Moreover, they excessively focus on local attributes and heavily rely on robust makeup generation, limiting adversarial transferability. 
In contrast, GIFT~\cite{gift} enhances transferability through global facial manipulation using GAN-inversion~\cite{ganinversion}, but it often introduces noticeable artifacts and unnatural modifications. In summary, it remains challenging for existing facial privacy protection methods to balance visual naturalness and adversarial transferability for practical applications effectively.

To address these challenges, we propose DiffAIM\footnote{The code is available at \url{https://github.com/Liqin-Wang/DiffAIM}}. Our approach leverages a pre-trained diffusion model~\cite{ddpm,ddim} to generate natural and highly transferable adversarial faces, without requiring extra reference information or retraining. DiffAIM manipulates facial identity within the diffusion model's latent space, leveraging its effective generative prior to produce more natural images. Specifically, we first employ edit-friendly latent mapping to map the source face image into a low-dimensional latent space. This step provides a good initialization for manipulation and the capability for faithful reconstruction. We then perform identity manipulation via iterative and gradient-based adversarial guidance injection during the reverse diffusion process. This iterative guidance allows for controllable and progressive steering toward the desired adversarial distribution. The guidance is derived from a carefully formulated objective for effective impersonation and visual naturalness. This objective promotes identity convergence towards a specific target, utilizing an adaptive ensemble of diverse FR models to enhance black-box transferability. Concurrently, it encourages semantic divergence from the source by utilizing intermediate features from the denoising U-Net~\cite{unet}, facilitating natural-looking impersonation. We further incorporate structure-preserving regularization to preserve facial structure during manipulation. Finally, an identity-sensitive timestep truncation strategy is introduced to balance adversarial effectiveness and visual quality. In summary, our main contributions are:
\begin{itemize}
    \item We propose DiffAIM for facial privacy protection, which manipulates facial identity by injecting gradient-based adversarial guidance into the reverse diffusion process, intending to craft natural adversarial faces that effectively deceive black-box FR systems. 

    \item We introduce a novel guidance objective that integrates FR-driven identity convergence with U-Net-driven semantic divergence, enabling natural impersonation. We further propose structure-preserving regularization to preserve facial structure during manipulation.

    \item Extensive experiments on face verification and identification tasks demonstrate that DiffAIM achieves state-of-the-art facial privacy protection performance against various deep FR models and commercial APIs, while maintaining superior visual quality.
\end{itemize}


\section{Related Works}
\subsection{Adversarial Attacks on Face Recognition}
Adversarial attacks~\cite{zhou2023downstream,zhou2023advclip,zhou2024securely} present a key strategy by introducing perturbations to input images during inference to mislead the FR systems. Early methods focused on noise-based attacks~\cite{noisebased_1} constrained by $l_p$ norms, but often produce perceptible artifacts.

\begin{figure}[t]
    \centering
    \includegraphics[width=\linewidth]
    {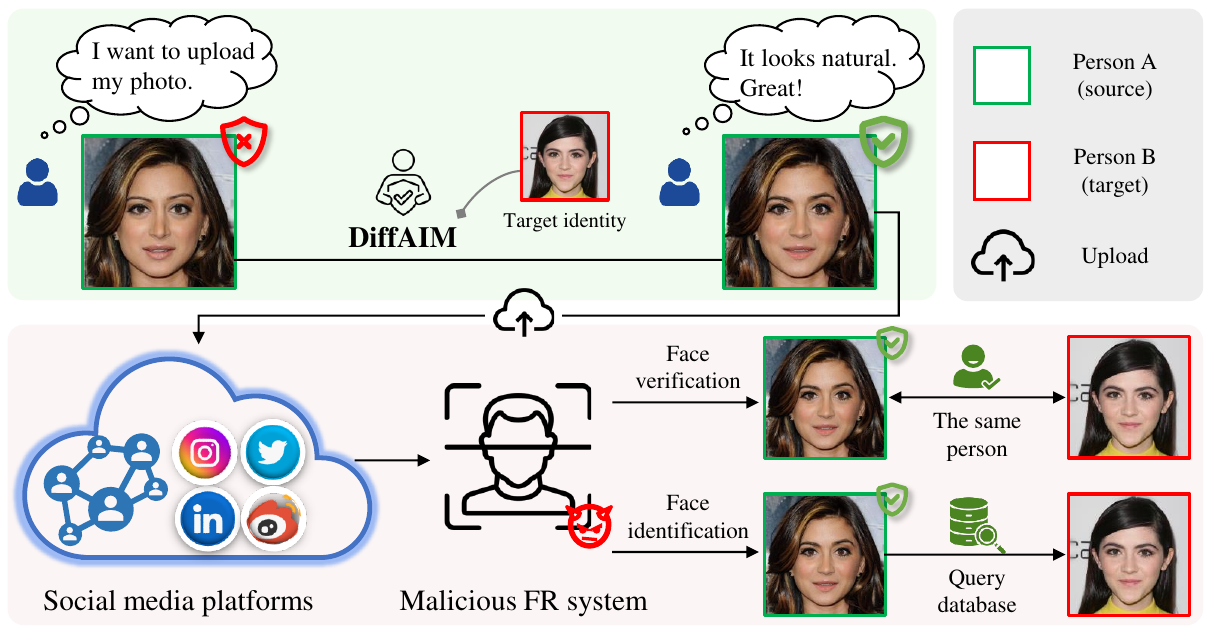}
    \caption{Illustration of our approach for facial privacy protection, which can generate a natural adversarial face image impersonating a specific target identity to deceive the malicious FR system in both face verification and face identification scenarios.}
    \label{fig:visual-intro}
\end{figure}
Another category of solutions employs unrestricted adversarial examples~\cite{unrestricted_2}, which are not constrained by the perturbation norm in pixel space, thereby improving image quality.
Within this category, makeup-based methods~\cite{advmakeup, amtgan, clip2protect} embed adversarial perturbations into synthesized facial makeup. However, these techniques often require external user-specified makeup reference information and can generate unnatural or heavy makeup appearances. Furthermore, their typical focus on local attributes and dependence on makeup generation can limit black-box transferability.
In contrast, Adv-Diffusion~\cite{adv-diffusion} utilizes the inpainting capabilities~\cite{inpainting_1} of diffusion models to generate higher-quality adversarial faces, but it exhibits limited attack transferability and destruction on the background content. GIFT~\cite{gift} employs global facial manipulation via GAN inversion~\cite{ganinversion} optimization, but it often introduces substantial and unrealistic alterations to the original face.
In this work, we propose DiffAIM to craft natural-looking adversarial faces without requiring extra reference information or model retraining.

\subsection{Diffusion Methods}
Diffusion models~\cite{ddpm,ddim,scorebased} are a class of probabilistic generative models renowned for their ability to synthesize high-quality images. The reverse process can be viewed as iterative denoising, transforming a simple prior distribution into a complex target distribution. To enhance computational efficiency, latent diffusion models (LDMs)~\cite{ldm} perform the diffusion process within a compressed latent space learned by an autoencoder~\cite{autoencoder_1,autoencoder_2}.

The powerful generative capabilities of diffusion models have led to their widespread adoption in various tasks, including text-to-image synthesis~\cite{t2i_1,t2i_2,ldm}, image editing~\cite{diffedit_1,diffedit_2,diffedit_3} and image super-resolution~\cite{diffsuper_1,diffsuper_2}. Further, recent research integrates adversarial example generation with the diffusion process to produce high-quality adversarial examples~\cite{diff_adv_1}. For instance, Advdiff~\cite{advdiff} uses a conditional diffusion model, adding adversarial perturbations during the denoising process starting from Gaussian noise. ACA~\cite{aca} modifies the initial latent code before commencing the reverse diffusion process to embed adversarial semantics. Inspired by these works, we utilize a latent diffusion model and introduce a fine-grained iterative guidance mechanism to manipulate facial identity, aiming to generate natural yet effective adversarial faces.
\section{Preliminaries}
\subsection{Problem Formulation}
FR systems generally operate in two modes: \textit{face verification} and \textit{face identification}. For face verification, the FR system determines if two face images belong to the same identity, often by comparing their feature similarity against a threshold. For face identification, the FR system queries the face database (gallery) to find the identity most similar to the input image (probe). An effective facial privacy protection approach should mislead the FR system in both scenarios.

To effectively protect facial privacy, as depicted in Fig.~\ref{fig:visual-intro}, given a source face image $I^{src}$, we aim to generate an adversarial face image $I^{adv}$ that satisfies: 1) \textbf{Naturalness}: it should not introduce perceptible artifacts or distortions and should preserve the human-perceived identity. 2) \textbf{Impersonation}: it should deceive the malicious FR system into identifying it as a specific target image $I^{tgt}$. For verification, this implies $Sim(\mathcal{M}(I^{adv}), \mathcal{M}(I^{tgt}))>\tau$, where $\mathcal{M}$ is the FR feature extractor, $Sim(\cdot, \cdot)$ is a feature-level similarity metric and $\tau$ is a threshold. For identification, we aim to let $Sim(\mathcal{M}(I^{adv}), \mathcal{M}(I^{tgt})) > Sim(\mathcal{M}(I^{adv}), \mathcal{M}(I^{src}))$. The problem we aim to solve can be formulated as
\begin{equation}\label{eq:adversarial_formulation_revised}
\begin{aligned}
    \max \limits_{I^{adv}} \ & [ Sim(\mathcal{M}(I^{adv}), \mathcal{M}(I^{tgt})) - Sim(\mathcal{M}(I^{adv}), \mathcal{M}(I^{src}))] \\
    & \text { s.t. } \mathcal{H}(I^{adv}, I^{src}) \leq \epsilon
\end{aligned}
\end{equation}
where $\mathcal{H}(I^{adv}, I^{src})$ measures the degree of unnaturalness of $I^{adv}$ in relation to $I^{src}$, with $\epsilon$ controlling the bound.

\begin{figure*}[t]
    \centering
    \includegraphics[width=\linewidth]{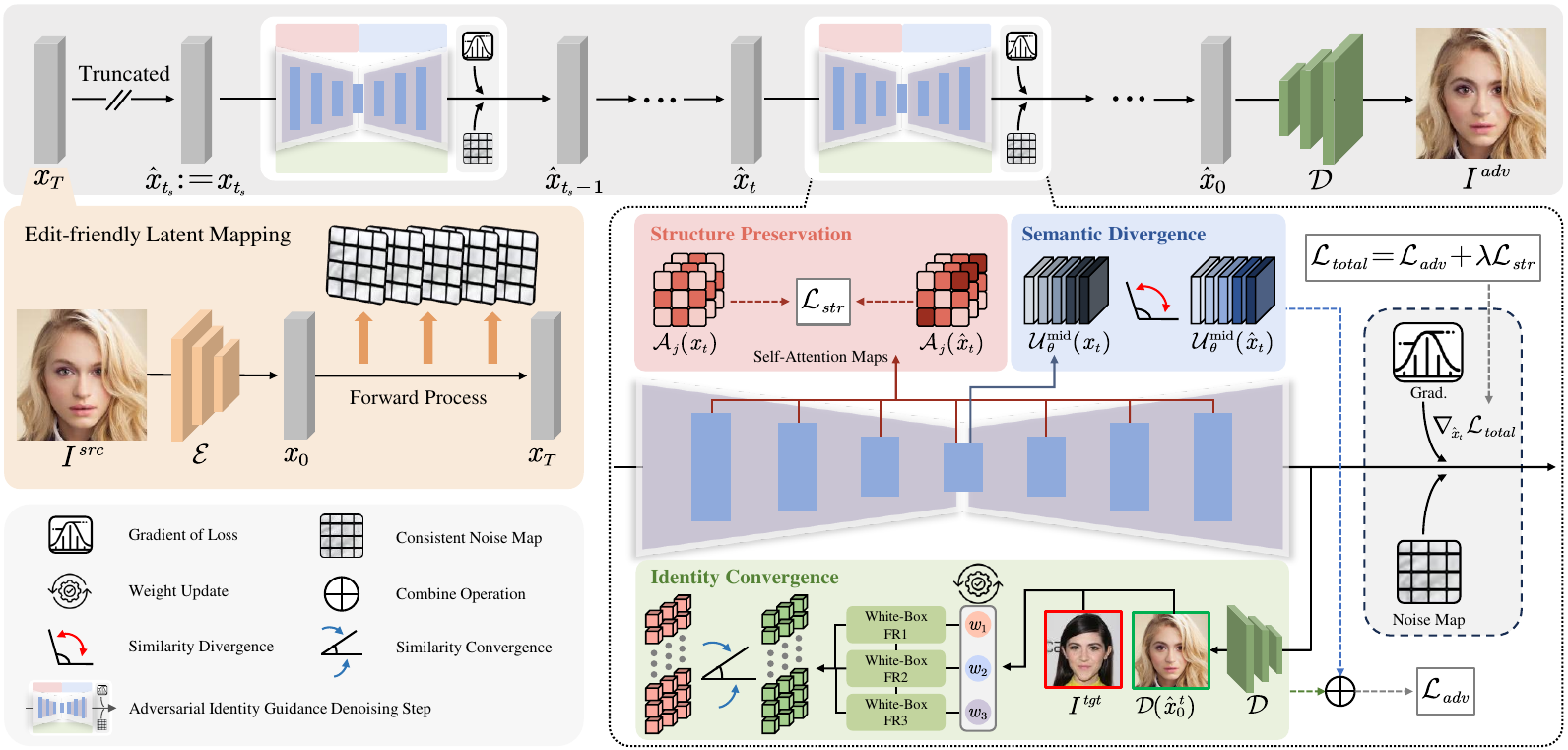}
    \caption{Overview of the proposed DiffAIM. We first employ edit-friendly latent mapping to obtain the initial latent code and consistent noise maps for reconstruction. Next, adversarial guided diffusion is used to generate natural adversarial face images.}
    \label{fig:visual-overview}
\end{figure*}

\subsection{Edit-Friendly Diffusion Inversion}
Diffusion inversion techniques~\cite{diff_inv_1, diff_inv_2} are widely used in diffusion-based image editing tasks. These techniques invert a real image into the latent space of a diffusion model, often by determining some initial noise that can be denoised to reconstruct the image. The noise can then be perturbed or modified during the generation process to edit the image. Recent work has shown that inverting the DDPM scheduler can achieve faithful reconstruction and few-step editing and propose the edit-friendly inversion~\cite{edit_fri}.

Concretely, recall the reverse diffusion process as follows
\begin{equation}\label{eq:reverse_process}
   x_{t-1} = \hat{\mu}_t(x_t) + \sigma_t z_t,\quad t = T, \ldots, 1
\end{equation}
where $\{z_t\}$ are iid standard normal vectors, $\sigma_t$ is derived from the noising schedule, and $\hat{\mu}_t(x_t)$ is derived from the output of denoising network, commonly implemented as a U-Net. 

Edit-friendly inversion aims to extract $\{ x_T, z_T, z_{T-1}, \ldots, z_1 \}$ from a given image $x_0$, which are then used in Eq.~\ref{eq:reverse_process} to reconstruct $x_0$. Specifically, given an image $x_0$, one first computes its noise sequence $\{x_t\}$ using independently sampled noise based on the standard forward diffusion process at each timestep
\begin{equation}\label{eq_edit-friendly_forward}
x_t = \sqrt{\bar{\alpha}_t} \, x_0 + \sqrt{1 - \bar{\alpha}_t} \, \tilde{\epsilon}_t, \quad t = 1, \ldots, T
\end{equation}
where, $\tilde{\epsilon}_t \sim \mathcal{N}(0, \mathbf{I})$ denotes the statistically independent noise. Given two such noisy images, $x_t$ and $x_{t-1}$, one then calculates the noise map $z_t$ that would be needed for Eq.~\ref{eq:reverse_process} to denoise $x_t$ to $x_{t-1}$
\begin{equation}\label{eq_edit-friendly_zt}
z_t = \frac{x_{t-1} - \hat{\mu}_t(x_t)}{\sigma_t}, \quad t = T, \ldots, 1
\end{equation}

Finally, one can simply denoise an image from $x_T$ with the pre-computed $\{z_t\}$ applied  at each timestep. In this work, we leverage this inversion technique as our latent mapping foundation, providing initialization and reconstruction capability for manipulation.

\section{Methodology}
\subsection{Overview}
We aim to generate adversarial face images that appear visually natural while effectively impersonating a target identity to deceive malicious FR systems. As illustrated in Fig.~\ref{fig:visual-overview}, our method achieves this by manipulating facial identity within the latent space of a pre-trained diffusion model, leveraging its powerful generative prior to produce natural images. The process begins with edit-friendly latent mapping to obtain editable latent codes from the source face for initialization, and consistent noise maps for faithful reconstruction. This step provides the necessary initialization and reconstruction capability for manipulation. Subsequently, we introduce adversarial guidance injection to refine the latent code at each timestep iteratively, enabling progressive steering toward the desired adversarial target distribution. Furthermore, we detail the adversarial identity guidance optimization, formulating and optimizing an objective function to derive the adversarial guidance. Finally, we introduce identity-sensitive timestep truncation, restricting guidance injection to later diffusion stages to effectively balance adversarial effectiveness and visual quality.

\subsection{Adversarial Identity Manipulation}
\label{sec:adv_manipulation}
The low-dimensional latent space of the diffusion model is well-trained on natural images and exhibits powerful generative capabilities~\cite{ldm}. Thus, we propose to manipulate identity within this latent space, utilizing its generative prior to craft natural adversarial faces. Specifically, we first map the source face into the latent space for initialization. Next, we iteratively guide the reverse diffusion process to achieve identity impersonation.

\textbf{Edit-friendly Latent Mapping.} 
To obtain the initial latent code suitable for manipulation, we employ the edit-friendly diffusion inversion technique~\cite{edit_fri} as our latent mapping foundation. This approach allows for faithful reconstruction from the inverted latent codes and noise maps, providing a solid foundation for subsequent manipulation without requiring complex inversion optimization.

Specifically, given a source face image $I^{src}$, we first encode it into the low-dimensional latent space using the pre-trained VAE encoder $\mathcal{E}$, obtaining $x_0 = \mathcal{E}(I^{src})$. We then compute the noisy latent codes $\{x_1, \ldots, x_T\}$ using the forward diffusion process with statistically independent noise at each timestep, as defined in Eq.~\ref{eq_edit-friendly_forward}. Concurrently, we derive the consistent noise maps $\{z_1, \ldots, z_T\}$ using Eq.~\ref{eq_edit-friendly_zt}. The sequence $\{x_t\}_{t=1}^T$ represents a benign diffusion trajectory originating from $x_0$. The consistent noise maps $\{z_t\}_{t=1}^T$ ensure we can reconstruct $x_0$ by initiating the reverse process from any $x_t$. We use the initial $x_T$, and the noise maps $\{z_t\}_{t=1}^T$ as inputs for the guided generation process.

\textbf{Adversarial Guidance Injection.}
To manipulate the identity, we inject adversarial guidance during the reverse diffusion process. Instead of only modifying the initial $x_T$, we iteratively refine the latent at each denoising step $t$. This allows for more fine-grained and controllable manipulation.

Specifically, we denote $\hat{x}_t$ as the adversarial latent code at timestep $t$. We first estimate the corresponding clean latent $\hat{x}^{t}_0$ from $\hat{x}_t$ as $\hat{x}^{t}_0 = (\hat{x}_t - \sqrt{1-\bar{\alpha}_t}\hat{\epsilon}_\theta(\hat{x}_t))/\sqrt{\bar{\alpha}_t}$. This predicted clean latent code is then decoded to obtain an intermediate adversarial image as $\mathcal{D}(\hat{x}^{t}_0)$, where $\mathcal{D}$ is the pre-trained VAE decoder. We input this intermediate image into the FR model $\mathcal{M}$ to compute the loss function $\mathcal{L}$ (detailed in the next section). The gradient of $\mathcal{L}$ with respect to the current adversarial latent $\hat{x}_t$ provides the adversarial guidance
\begin{equation}
    \mathcal{G}_t = \nabla_{\hat{x}_t}\mathcal{L}(\mathcal{D}(\hat{x}^{t}_0), I^{tgt}, \mathcal{M})
\end{equation}

The update rule combines the standard reverse diffusion step using pre-computed $z_t$ with the adversarial guidance $\mathcal{G}_t$. The adversarial latent $\hat{x}_{t-1}$ for the previous timestep is computed as
\begin{equation}\label{eq_adversarial_latent_update}
\hat{x}_{t-1} = \hat{\mu}_t(\hat{x}_t) + \sigma_t z_t + \mathcal{G}_t
\end{equation}

By iteratively applying adversarial guidance injection from $t=T$ down to $t=1$, the final adversarial face image $I^{adv}$ can be obtained by decoding the resulting latent code as $I^{adv}=\mathcal{D}(\hat{x}_0)$. 

\subsection{Adversarial Identity Guidance Optimization}
\label{sec:adv_opt}
We now detail the formulation and optimization of the objective function used to derive the adversarial guidance, aiming to steer the generation towards natural impersonation.

\textbf{Adversarial Loss.}
The primary objective is to ensure effective impersonation. A naive way, aligning with the objective in Eq.~\ref{eq:adversarial_formulation_revised}, might directly optimize for identity convergence to $I^{tgt}$ and identity divergence from $I^{src}$ using the FR model
\begin{equation}\label{eq_naive_loss}
    \mathcal{L}_{adv}^{naive} = Sim(\mathcal{M}(\mathcal{D}(\hat{x}^{t}_0)), \mathcal{M}(I^{tgt})) - Sim(\mathcal{M}(\mathcal{D}(\hat{x}^{t}_0)), \mathcal{M}(I^{src}))
\end{equation}
where $Sim(v_1,v_2)=\cos(v_1,v_2)$. However, we empirically observed that directly optimizing the FR-driven divergence term often introduces perceptible visual artifacts, degrading image quality.

Recent work~\cite{unet_hspace} has demonstrated that manipulating the deepest features within the intermediate U-Net block can lead to meaningful semantic edits without degrading image quality. Inspired by this, we hypothesize that leveraging high-level semantic features from U-Net for the divergence component can effectively push the generated image away from the source without causing significant visual degradation, thereby facilitating natural-looking impersonation. Consequently, we propose a refined adversarial loss that combines a FR-driven identity convergence term with a U-Net-driven semantic divergence term
\begin{equation}\label{eq_improved_adv_loss}
    \mathcal{L}_{adv} = \underbrace{-Sim(\mathcal{U}_{\theta}^{\text{mid}}(\hat{x}_t), \mathcal{U}_{\theta}^{\text{mid}}(x_t))}_{\text{semantic divergence}} +
    \underbrace{Sim(\mathcal{M}(\mathcal{D}(\hat{x}^{t}_0)), \mathcal{M}(I^{tgt}))}_{\text{identity convergence}}
\end{equation}
where $x_t$ is the benign inverted latent code produced in the latent mapping process, and $\hat{x}_t$ is the corresponding adversarial latent code, $\mathcal{U}_{\theta}^{\text{mid}}(\cdot)$ represents the features output from the intermediate block of the U-Net. This formulation combines the explicit discriminative power of FR models with the rich semantic and generative priors learned by U-Net, leading to more effective adversarial results and better visual naturalness compared to the naive ones.

\begin{algorithm}[t]
\SetKwInput{Para}{\textbf{Parameter}}
\caption{The algorithm of DiffAIM}
\label{alg}
\KwIn{Source image ${I}^{src}$, target image ${I}^{tgt}$, pre-trained LDM, diffusion steps $T$, truncated timestep ${t}_{s}$, attack iterations ${N}_{a}$, step size $\eta$. }
\KwOut{Adversarial face image ${I}^{adv}$.}
Generate initial latent code ${x}_{0}=\mathcal{E}({I}^{src})$\;
Calculate benign latent codes $\{x_1, \ldots, x_T\}$ and consistent noise maps $\{z_1, \ldots, z_T \}$ using Edit-friendly Latent Mapping\;
Initialize $\{w_i\}_{i=1}^{N_m}=1/N_m$, let $\hat{x}_{t_s} = x_{t_s}$\;
\For{$t = {t}_{s}$ down to $1$}
{
    ${x}_{t-1} = {\mu}_t(\hat{x}_t)+\sigma_tz_t, \ \mathcal{G}_t^0 = 0$\;
    \For{$k = 1$ to $N_a$}
    {
        $\hat{x}^{t}_0 = \frac{\hat{x}_t - \sqrt{1-\bar{\alpha}_t}\hat{\epsilon}_\theta(\hat{x}_t)}{\sqrt{\bar{\alpha}_t}}$\;
        
        Calculate $\mathcal{L}_{adv} ,\ \mathcal{L}_{str} ,\ \mathcal{L}_{total}$\;
        
        $w_i \leftarrow \text{Update}(\mathcal{M}_i,\mathcal{D},\hat{x}^{t}_0, I^{tgt})$\;
        
        $\hat{x}_t = \hat{x}_t+\Pi_{\kappa} ( \mathcal{G}_t^{k-1} + \eta \cdot \nabla_{\hat{x}_t}\mathcal{L}_{total})$\;
    }
    $\hat{x}_{t-1}=x_{t-1}+ \mathcal{G}_t^{N_a}$\;
}
\textbf{Return the adversarial face image $I^{adv} = \mathcal{D}(\hat{x}_0) $.}
\end{algorithm}

\textbf{Adaptive Ensemble Strategy.}
To further enhance the transferability of the generated adversarial faces against unseen black-box FR systems, we select $N_m$ pre-trained FR  models $\{\mathcal{M}_i\}_{i=1}^{N_m}$ with high accuracy on public face datasets, serving as white-box models to simulate the decision boundaries of potential target models under the black-box setting. The final form of adversarial loss is
\begin{equation}\label{eq_attack_loss}
    \mathcal{L}_{adv} = 
    - Sim(\mathcal{U}_{\theta}^{\text{mid}}(\hat{x}_t), \mathcal{U}_{\theta}^{\text{mid}}(x_t)) +
    \sum_{i=1}^{N_m} w_i \cdot Score_i 
\end{equation}
where $\{w_i\}$ is a set of weights determined by the robustness of each white-box FR model. $Score_i=Sim(\mathcal{M}_i(\mathcal{D}(\hat{x}^{t}_0))), \mathcal{M}_i(I^{tgt}))$. We would like to make the harder-to-learn task, i.e., the harder-to-attack FR model, has higher weights during optimization. Hence, we use the similarity score to adaptively update the weight of each white-box FR model in the following way
\begin{equation}\label{eq_weight_update}
   w_i = \frac{e^{1-Score_i}}{\sum_{i=1}^{N_m}e^{1-Score_i}}
\end{equation}
Initially, $\{w_i\}_{i=1}^{N_m}$ is set to $1/N_m$ and adaptively updated throughout the iterative adversarial identity manipulation process. 

\textbf{Structural Preservation Loss.}
Adversarial manipulation at each timestep can inadvertently distort the underlying facial structure. Inspired by findings~\cite{selfattn} that self-attention layers can capture structure information within images, we further introduce regularization to maintain structural consistency between the adversarial latent $\hat{x}_t$ and the benign latent $x_t$. We minimize the difference between their respective self-attention maps as
\begin{equation}\label{eq_struecture_loss}
    \mathcal{L}_{str} = -\sum_{j\in \mathcal{S}} 
    \lvert \lvert \mathcal{A}_j(\hat{x}_t)-\mathcal{A}_j(x_t) \lvert \lvert^2_2
\end{equation}
where $\mathcal{S}$ is the set of indices for self-attention layers in the U-Net, and $\mathcal{A}_j(\cdot)$ extracts the $j$-th self-attention map. This encourages the adversarial latent to retain structural characteristics similar to the benign latent at the same diffusion stage, thereby preserving the facial structure of the generated image.

\textbf{Total Loss and Optimization.}
By combining all the above loss, we have the total loss function as
\begin{equation}\label{eq_total_loss}
    \mathcal{L}_{total}= 
    \mathcal{L}_{adv} + \lambda\mathcal{L}_{str}
\end{equation}
where $\lambda$ represent the hyper-parameter. 
\begin{figure}[t]
    \centering
    \includegraphics[width=\linewidth]{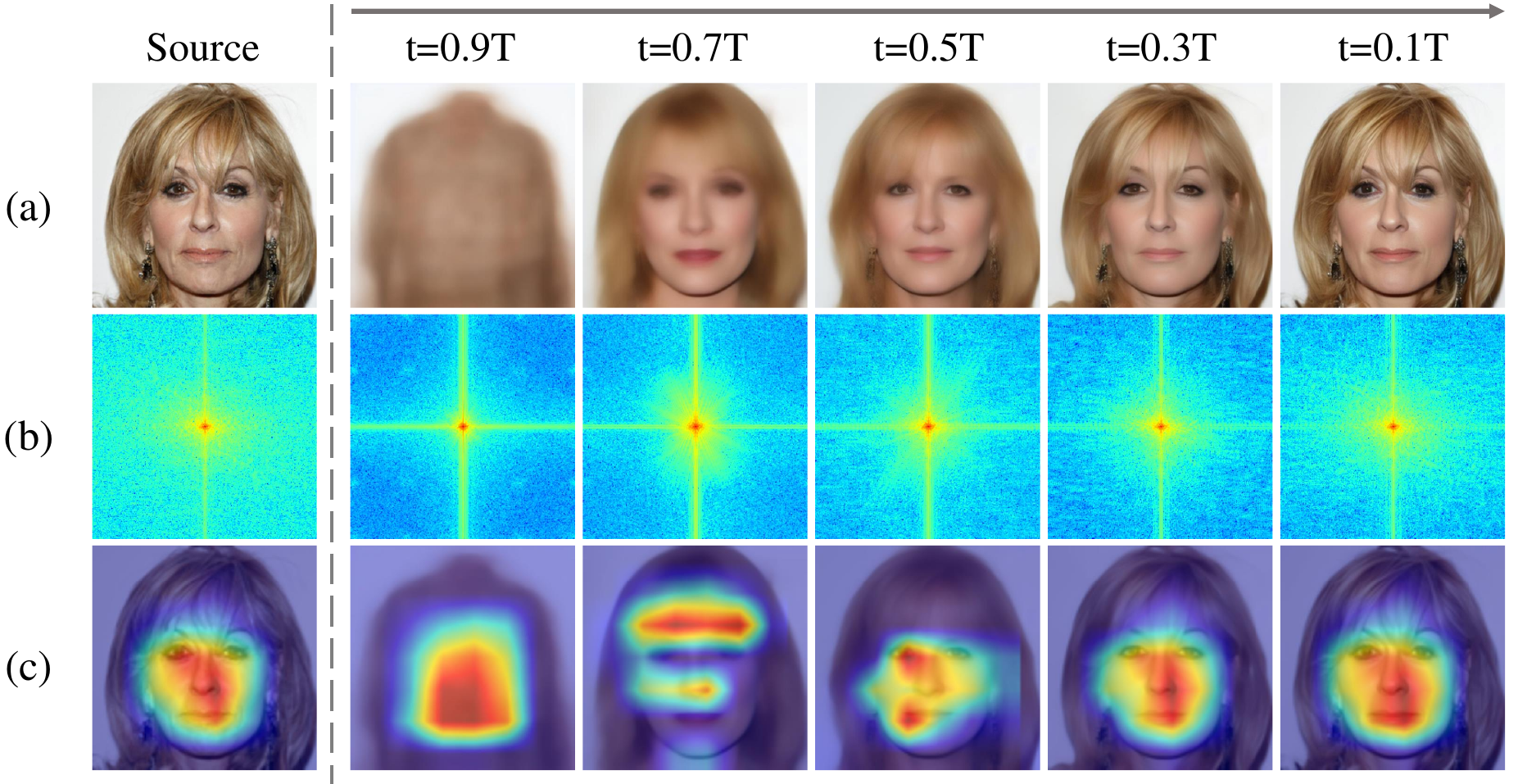}
    \caption{(a) Final output predicted at different timestep. (b) The spectrograms of the predicted final output. (c) Visualization of gradient response using Grad-CAM on the FR model.}
    \label{fig:visual-timestep_explore}
\end{figure}
In practice, we employ an iterative refinement procedure to obtain more stable and effective adversarial guidance $\mathcal{G}_t$. At each timestep $t$, we perform $N_a$ inner optimization steps
\begin{equation}
    \mathcal{G}_{t}^{k} = \Pi_{\kappa} ( \mathcal{G}_{t}^{k-1} + \eta \cdot \nabla_{\hat{x}_t}\mathcal{L}_{total}), \quad k=1, \ldots, N_a
\end{equation}
where $\eta$ is the step size, and $\Pi_{\kappa}$ is defined as the projection of the adversarial guidance onto $\kappa$-ball. After $N_a$ iterations, we can obtain $\mathcal{G}_t = \mathcal{G}_{t}^{N_a}$, which is then used in adversarial latent update (Eq.~\ref{eq_adversarial_latent_update}).

\begin{figure*}[t]
    \centering
    \includegraphics[width=0.75\textwidth]{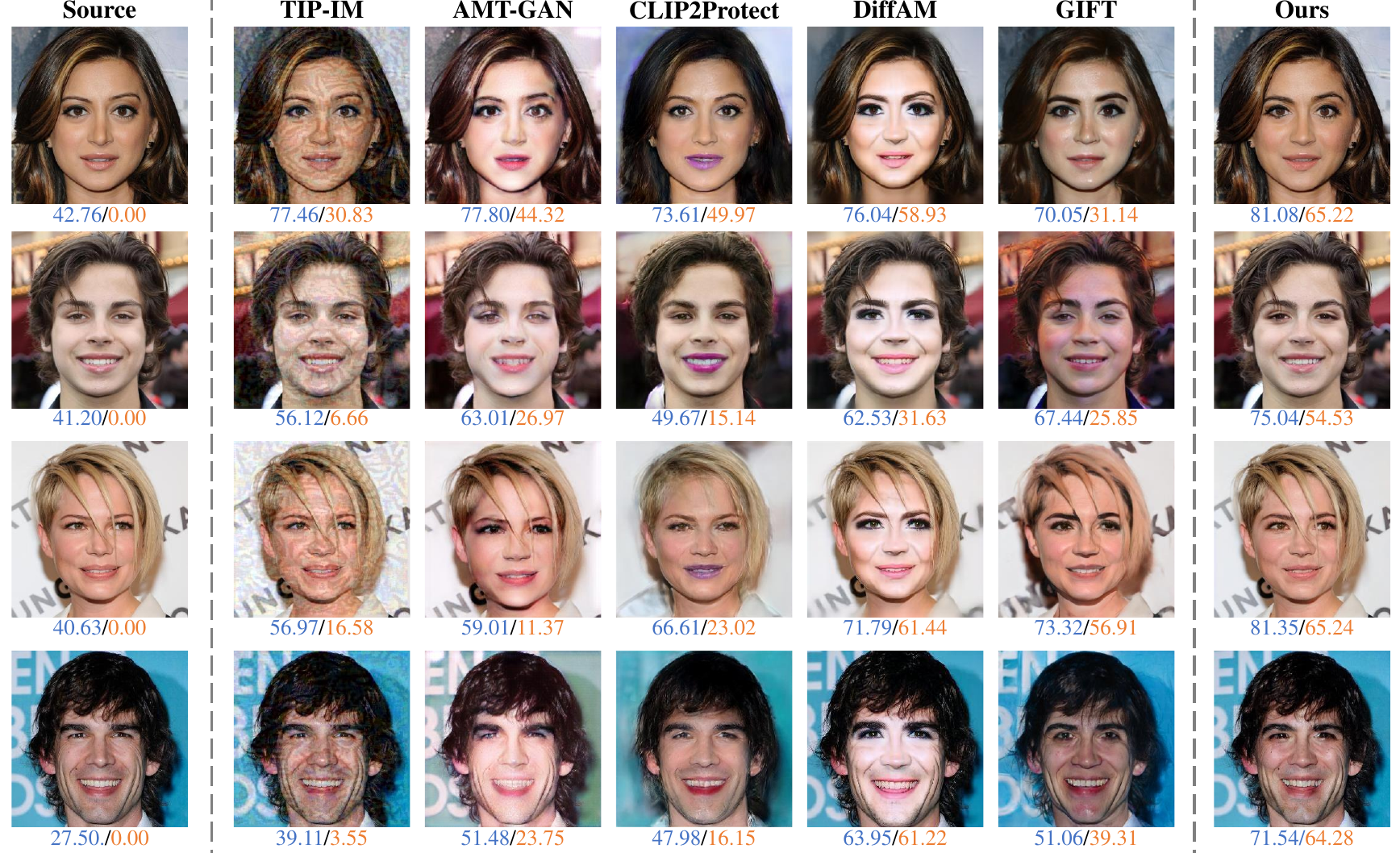}
    \caption{Visualizations of the adversarial face images generated by different facial privacy protection methods. The numbers below each image are confidence scores returned by Face++ and Aliyun.}
    \label{fig:visual-compare}
\end{figure*}

\begin{table*}[t]
    \begin{center}
    \caption{Attack success rate (\%) of impersonation attack under the face verification task.}
    \setlength{\tabcolsep}{8.0pt}
    \resizebox{0.85\textwidth}{!}
    {
    \begin{tabular}{l || c c c c || c c c c || c }
    \toprule[0.13em]
    \textbf{Methods} & \multicolumn{4}{c||}{\textbf{CelebA-HQ}}&\multicolumn{4}{c||}{\textbf{LADN-Dataset}}&\multicolumn{1}{c}{\textbf{Average}} \\
    & IRSE50 & IR152 & FaceNet& MobileFace & IRSE50 & IR152 & FaceNet& MobileFace &  \\
    \midrule[0.13em]
     Clean & 7.29 & 3.80 & 1.08 & 12.68 & 2.71 & 3.61 & 0.60 & 5.11 & 4.61 \\
     PGD & 36.87 & 20.68 & 1.85 & 43.99 & 40.09 & 19.59 & 3.82 & 41.09 & 26.00 \\
     TIP-IM & 54.40 & 37.23 & 40.74 & 48.72 & 65.89 & 43.57 & 63.50 & 46.48 & 50.07 \\
     Adv-Makeup & 21.95 & 9.48 & 1.37 & 22.00 & 29.64 & 10.03 & 0.97 & 22.38 & 14.73 \\
     AMT-GAN & 76.96 & 35.13 & 16.62 & 50.71 & 89.64 & 49.12 & 32.13 & 72.43 & 52.84 \\
     CLIP2Protect & 81.10 & 48.42 & 41.72 & 75.26 & 91.57 & 53.31 & 47.91 & 79.94 & 64.90 \\
     DiffAM & 92.00 & 63.13 & 64.67 & 83.35 & \textbf{95.66} & 66.75 & 65.44 & \textbf{92.04} & 77.88 \\
     GIFT & 95.20 & 75.50 & 65.30 & 92.20 & 92.17 & 76.20 & 69.88 & 83.73 & 81.27 \\
    \midrule
     Ours & \textbf{98.30} & \textbf{84.90} & \textbf{79.70} & \textbf{96.90} & 94.88 & \textbf{76.21} & \textbf{71.69} & 86.45 & \textbf{86.13} \\
    \bottomrule[0.1em]
    \end{tabular}
    }
    \label{table:verification_asr}
    \end{center}
\end{table*}

\subsection{Identity-Sensitive Timestep Truncation}
\label{sec:timestep_trunc}
We find that injecting adversarial guidance throughout the reverse diffusion process often leads to severe artifacts and distortions in the generated face images. Therefore, we aim to explore a timestep strategy to balance adversarial effectiveness and visual quality.

We begin with the fact that diffusion models typically establish low-frequency global facial structures in earlier steps before refining high-frequency details concentrated in identity-sensitive regions (e.g., eyes, nose, cheek, Fig.~\ref{fig:visual-timestep_explore} (a) and (b)), which cognitive psychology identifies as vital for identity discrimination. Second, we analyzed the attention patterns of the FR model across timestep using Grad-CAM~\cite{gradcam}, as shown in Fig.~\ref{fig:visual-timestep_explore} (c). This revealed that the model's attention gradually shifts from a dispersed pattern to a more focused one, concentrating on these key identity-sensitive regions during an intermediate phase of the reverse diffusion process, approximately from $0.5T$ down to $0.1T$.

These suggest that there exists a later phase, where the diffusion model actively refines identity-sensitive features, and the FR model is simultaneously highly sensitive to these features. We hypothesize that delaying the injection of adversarial guidance until this phase could enable effective adversarial manipulation, minimizing disruption to the previously established global facial structure and preserving visual quality.

Motivated by these insights, we propose a timestep truncation strategy wherein adversarial guidance is injected only from $t=t_s$ down to $t=1$, where $t_s$ is the carefully selected truncated timestep and typically $t_s \ll T$. Our ablation study validates this strategy, showing that selecting $t_s$ within the range $\left[0.2T, 0.3T\right]$ achieves a strong balance between adversarial effectiveness and visual quality. The algorithm of DiffAIM is presented in Algorithm~\ref{alg}.

\section{Experiments}
\subsection{Experimental Setup}
\textbf{Implementation Details:}
We employ stable diffusion~\cite{ldm} as the pre-trained diffusion model to generate adversarial face images. We utilize the DDIM sampler~\cite{ddim} with $T=100$ diffusion steps. We set the attack iterations $N_a=10$, the step size $\eta=3$, the truncated timestep $t_s=20$, and the hyper-parameter for structure preservation regularization is set to $\lambda=0.1$.

\textbf{Datasets:}
We conduct experiments for both face verification and identification tasks. \textit{Face Verification}: we use CelebA-HQ~\cite{celebahq} and LADN~\cite{ladn} as our test sets. Specifically, we select a subset of 1000 images with different identities from CelebA-HQ and divide them into 4 groups, with each group of images impersonating a target identity provided by~\cite{amtgan}. Similarly, for LADN, we divide the 332 images into 4 groups for attack on different target identities. \textit{Face Identification}: we use 500 randomly selected images with different identities from CelebA-HQ as the probe set. The gallery set is constructed using the corresponding 500 images of the same identities along with the 4 target identities provided by~\cite{amtgan}.

\textbf{Baseline Methods:}
We compare the proposed approach with multiple noise-based and unrestricted-based facial privacy protection methods.
Noise-based methods include PGD~\cite{pgd}, and TIP-IM~\cite{tipim}. Unrestricted-based methods include Adv-Makeup~\cite{advmakeup}, AMT-GAN~\cite{amtgan}, CLIP2Protect~\cite{clip2protect}, DiffAM~\cite{sun2024diffam}, and GIFT~\cite{gift}.

\textbf{Target Models:}
We aim to protect facial privacy by attacking deep FR models in the block-box settings. Specifically, our study targets widely used FR models, including IRSE50~\cite{irse50}, IR152~\cite{ir152}, FaceNet~\cite{facenet}, and MobileFace~\cite{mobilefacenets}. Meanwhile, we evaluate the performance on commercial FR APIs, including Face++ and Aliyun. 

\begin{table*}[t]
    \begin{center}
    \caption{Attack success rate (\%) of impersonation attack under the face identification task.}
    \setlength{\tabcolsep}{12.0pt}
    \resizebox{0.85\textwidth}{!}
    {
    \begin{tabular}{l || c c || c c || c c || c c || c c }
    \toprule[0.13em]
    \textbf{Methods} & \multicolumn{2}{c||}{\textbf{IRSE50}}&\multicolumn{2}{c||}{\textbf{IR152}}&\multicolumn{2}{c||}{\textbf{FaceNet}}&\multicolumn{2}{c||}{\textbf{MobileFace}}&\multicolumn{2}{c}{\textbf{Average}} \\
    & R1-T & R5-T & R1-T & R5-T & R1-T & R5-T & R1-T & R5-T & R1-T & R5-T  \\
    \midrule[0.13em]
    TIP-IM & 16.2 & 51.4 & 21.2 & 56.0 & 8.1 & 35.8 & 9.6 & 24.0 & 13.8 & 41.8 \\
    CLIP2Protect & 24.5 & 64.7 & 24.2 & 65.2 & 12.5 & 38.7 & 11.8 & 28.2 & 18.3 & 49.2 \\
    DiffAM & 41.2 & 85.4 & 21.8 & 41.2 & 34.8 & 60.8 & 13.6 & 52.2 & 27.9 & 59.9 \\
    GIFT & 27.2 & 89.0 & \textbf{29.0} & 64.2 & 22.8 & 53.6 & 19.8 & 72.0 & 24.7 & 69.7 \\
    \midrule
    Ours & \textbf{43.6} & \textbf{98.6} & 24.6 & \textbf{74.6} & \textbf{49.8} & \textbf{81.8} & \textbf{21.8} & \textbf{82.0} & \textbf{35.0} & \textbf{84.2} \\
    \bottomrule[0.1em]
    \end{tabular}}
    \label{table:identification_asr}
    \end{center}
\end{table*}

\begin{table}[t]
    \caption{Quantitative evaluations of image quality.}
    \setlength{\tabcolsep}{13.5pt}
    \resizebox{0.43\textwidth}{!}
    {
    \begin{tabular}{l | c  |c | c }
    \toprule[0.13em]
    \textbf{Methods} & PSNR $\uparrow$  & SSIM $\uparrow$ & FID $\downarrow$\\
    \midrule[0.13em]
    TIP-IM & 26.2564 & 0.7371 & 96.9909 \\
    AMT-GAN & 19.7968 & 0.7883 & 43.8369 \\
    CLIP2Protect  & 19.5175 & 0.6445 & 54.0374 \\
    DiffAM & 15.3209 & 0.5466 & 71.4500 \\
    GIFT& 18.7171 & 0.6586 & 46.2044 \\
    \midrule
    Ours (w/o $\mathcal{L}_{str}$) & 26.2516 & 0.7991 & 22.2159 \\
    Ours & \textbf{27.6814} & \textbf{0.8108} & \textbf{15.5600} \\
    \bottomrule[0.1em]\end{tabular}}
    \label{table:quantitative_visual}
\end{table}

\textbf{Evaluation Metrics:}
We use different evaluation strategies to calculate the attack success rate (ASR) for face verification and identification scenarios. For face verification, we compute ASR as the proportion of adversarial face images successfully misclassified by the malicious FR system. For face identification, we adopt the closed-set protocol and evaluate the Rank-N targeted identity success rate (Rank-N-T), where all gallery images are ranked in descending order of their similarity to the probe image. The attack is deemed successful if the target identity appears in the top N candidates. For commercial FR APIs, we directly record the confidence score returned by the respective servers. We also use PSNR (dB), SSIM~\cite{ssim}, and FID~\cite{fid} to evaluate image quality.

\subsection{Comparison Study}
\textbf{Comparison on Black-box FR Models.}
We present experimental results of DiffAIM on two public datasets against four pre-trained FR models in the black-box settings. To generate adversarial face images, we use three FR models as white-box models to simulate the decision boundaries of the fourth model. For face verification, 
we set the threshold at 0.01 false acceptance rate for each FR model, i.e., IRSE50 (0.241), IR152 (0.167), FaceNet (0.409), and MobileFace (0.302). As shown in Tab.~\ref{table:verification_asr}, our approach achieves the highest average attacking performance compared with the SOTA methods under the face verification task.
We also provide ASR under the face identification task in Tab.~\ref{table:identification_asr}, where our method achieves an average absolute gain of about 7\% and 15\% over other SOTA methods in Rank-1 and Rank-5 settings, respectively. The results show that DiffAIM has strong black-box transferability, demonstrating the role of adversarial identity manipulation as we expected.

\textbf{Comparison on Image Quality.}
Tab.~\ref{table:quantitative_visual} reports the evaluations of image quality. Our method consistently achieves superior performance across all image quality metrics, indicating that the adversarial face images generated by DiffAIM are more natural and have fewer artifacts and distortion.

We further present the qualitative comparison of visual quality in Fig.~\ref{fig:visual-compare}. 
As illustrated, TIP-IM introduces noticeable noise, resulting in clearly detectable modifications. AMT-GAN struggles to accurately align the generated makeup with facial characteristics and, leading to artifacts on faces. CLIP2Protect frequently fails to generate makeup corresponding to the textual prompt and tends to damage the background content. The adversarial faces generated by DiffAM have an obvious "painted-on" appearance that looks unnatural. 
\begin{table}
    \caption{Impact of the semantic divergence term (sem\_div).}
    \setlength{\tabcolsep}{10.0pt}
    \resizebox{0.43\textwidth}{!}
    {
    \begin{tabular}{ l | c c | c c }
    \toprule[0.15em]
    & \multicolumn{2}{c|}{\textbf{ Face Verification}} & \multicolumn{2}{c}{\textbf{Face Identification}}\\
     & CelebA-HQ & LADN & R1-T & R5-T \\
    \midrule[0.15em]
    w/o sem\_div & 77.30 & 70.18 & 45.4 & 79.4 \\
    w/ sem\_div & 79.70 & 71.69 & 49.8 & 81.8 \\
    \bottomrule[0.1em]\end{tabular}}
    \label{table:robust_triplet}
\end{table}
\begin{figure}[!t]
    \centering
    \includegraphics[width=\linewidth]{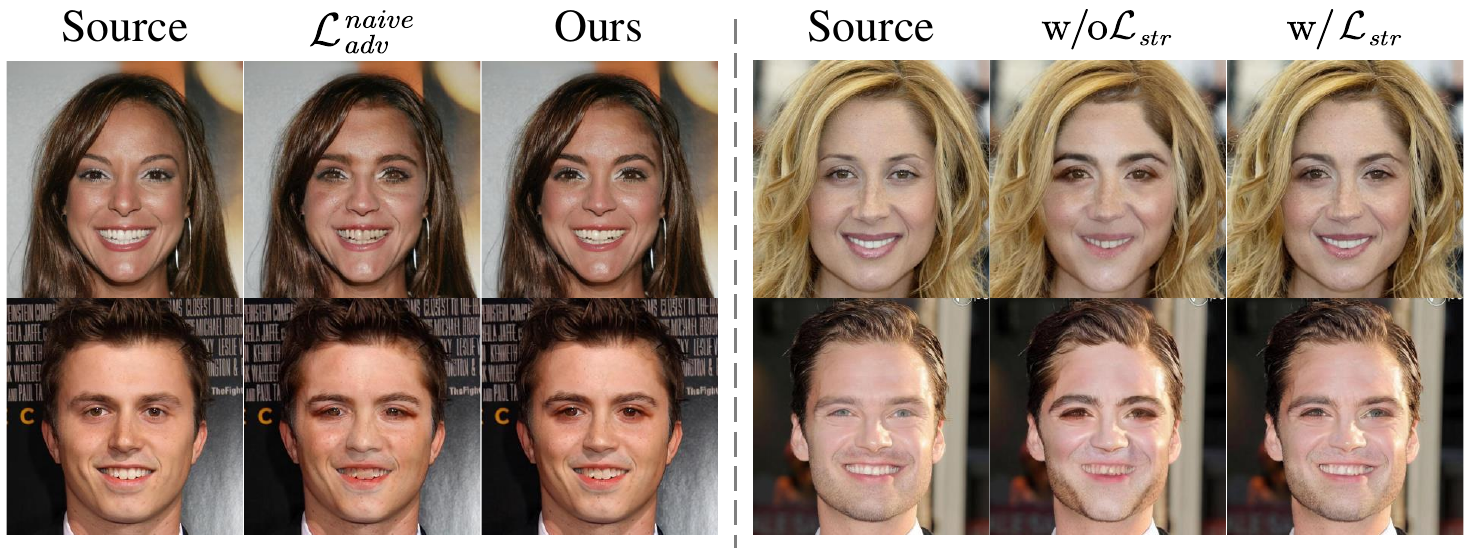}
    \caption{Ablation study for proposed the adversarial loss and structural preservation-regularization.}
    \label{fig:ablation_loss}
\end{figure}
Additionally, GIFT suffers from instability in the generation process, introducing uncontrollable modifications and noticeable artifacts. In contrast, we precisely manipulate on identity-relevant facial regions to generate natural-looking adversarial faces, while effectively preserving background content.  

\begin{figure*}[t]
    \centering
    \includegraphics[width=0.95\linewidth]{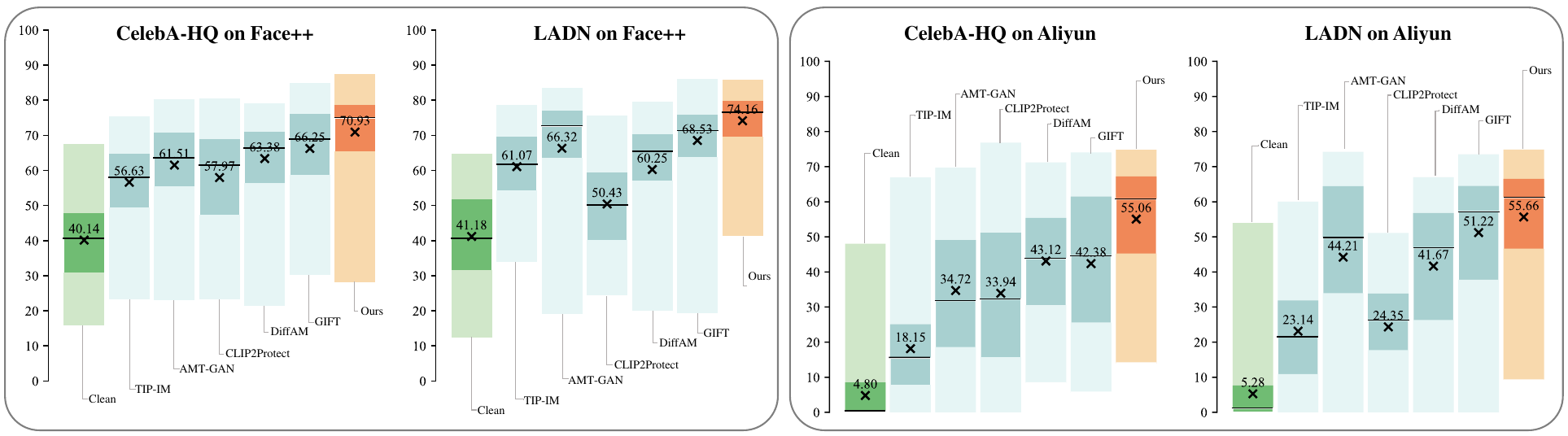}
    \caption{Confidence scores (higher is better) returned by Face++ and Aliyun. The number on each bar represents the average confidence score. Our approach has higher and more stable confidence scores than other facial privacy protection methods.}
    \label{fig:visual-api}
\end{figure*}

\begin{figure}[t]
    \centering
    \includegraphics[width=0.9\linewidth]{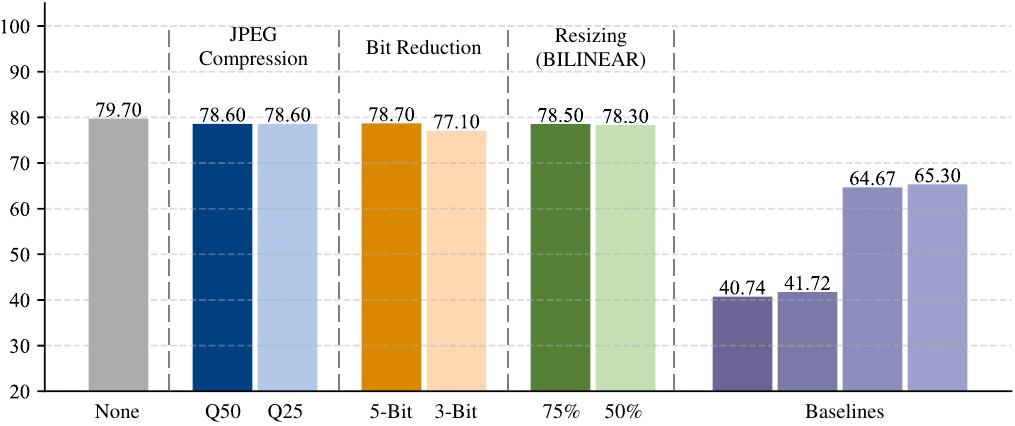}
    \caption{Robustness on various lossy operations. We report ASR (\%) on the FR model (FaceNet) on CelebA-HQ dataset under the face verification task. Baseline methods from left to right are TIP-IM, CLIP2Protect, DiffAM, and GIFT.}
    \label{fig:visual-robust}
\end{figure}

\textbf{Comparison on Commercial APIs.}
We further validate DiffAIM against commercial APIs such as Face++ and Aliyun in face verification mode. These APIs return confidence scores ranging from 0 to 100 indicating the similarity of identity between the adversarial image and the target image. Since the training data and model parameters of these commercial APIs are proprietary and undisclosed, it effectively simulates real-world facial privacy protection scenarios. We use several baseline methods and the proposed approach to protect 100 randomly selected images each from CelebA-HQ and LADN datasets, and record the confidence scores returned by these APIs. As shown in Fig.~\ref{fig:visual-api}, the results indicate that our method achieves the highest average confidence scores of about 72 and 55 on each API. The stable adversarial effectiveness across different datasets further underscores its practical applicability in real-world scenarios.

\subsection{Robust Study}\label{sec:ablation}
In real-world scenarios, images uploaded to social networks often undergo various lossy operations that can affect the effectiveness of adversarial face images. We thus evaluate the robustness of DiffAIM against common lossy operations, including JPEG compression, bit reduction, and resizing. As shown in Fig.~\ref{fig:visual-robust}, our method causes only minor performance degradation across various lossy operations, yet still demonstrates superior performance compared to other baselines, highlighting its significant robustness due to the semantically meaningful manipulation within the diffusion model's latent space.

\subsection{Ablation Study}
\textbf{Effect of Improved Adversarial Loss.}
As shown in Fig.\ref{fig:ablation_loss}, the naive adversarial loss introduces perceptible artifacts, particularly around the eyes and mouth. Conversely, our proposed refined adversarial loss generates adversarial faces with a more natural appearance. Additionally, as shown in Tab.~\ref{table:robust_triplet}, integrating the U-Net-driven semantic divergence achieves enhanced adversarial effectiveness.


\textbf{Effect of Structural Preservation.} 
Here, we demonstrate the importance of structural-preservation regularization in maintaining structural consistency and image quality. As shown in Fig.~\ref{fig:ablation_loss} and Tab.~\ref{table:quantitative_visual}, in the absence of the structural-preservation regularization, the generated adversarial faces exhibit distortions and drop in image quality. In contrast, the faces generated with structural preservation maintain better consistency in facial features and structure.

\begin{figure}[t]
    \centering
    \includegraphics[width=0.95\linewidth]{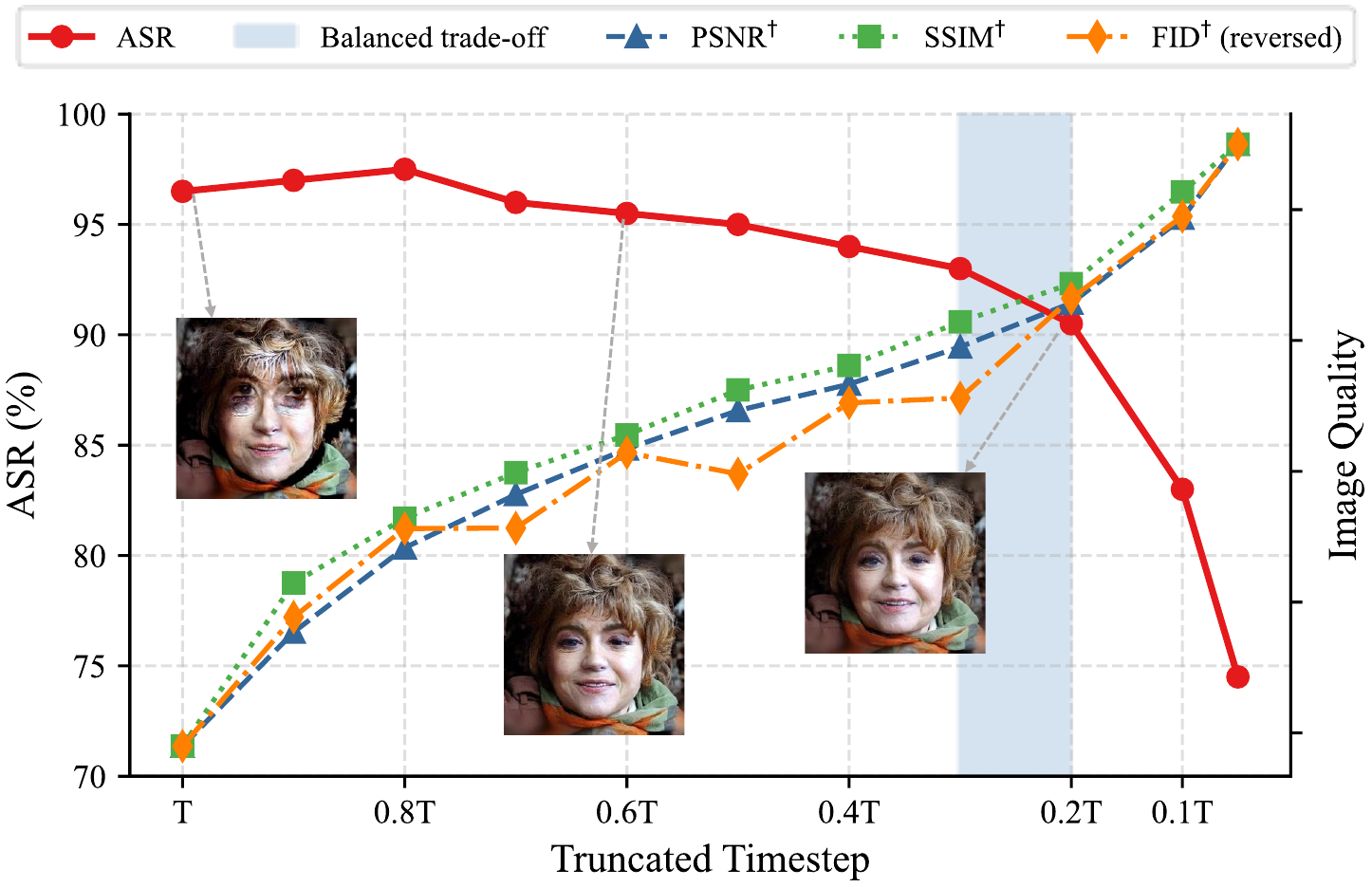}
    \caption{Ablation study for the truncated timestep $t_s$. Image quality metrics ($\dagger$) are scaled for visual trend comparison.}
    \label{fig:ablation_truncation}
\end{figure}
\textbf{Selection of truncated timestep.}
We analyze the impact of the truncated timestep $t_s$. As shown in Fig.~\ref{fig:ablation_truncation}, injecting adversarial guidance early yields higher ASR but degrades image quality. Specifically,  injecting guidance throughout the reverse diffusion process (i.e., $t_s=T$) introduces noticeable artifacts. Conversely, delaying guidance improves visual quality but causes a decline in ASR. The results reveal an optimal range around $t_s \in [0.2T, 0.3T]$, where ASR remains high while achieving excellent visual quality. 
\section{Conclusion}
In this paper, we introduce DiffAIM, a novel approach that leverages a pre-trained diffusion mode to generate natural adversarial faces. Building upon the generative capabilities of diffusion models, we manipulate facial identity by injecting gradient-based adversarial guidance into the reverse diffusion process. Our guidance is designed for identity convergence and semantic divergence, enabling natural impersonation. Moreover, we propose a structure-preserving regularization to preserve facial structure.
Extensive experiments show that DiffAIM is highly effective against different open-source FR models and commercial APIs while maintaining the superior visual quality. We believe DiffAIM offers a promising direction for practical facial privacy protection.



\begin{acks}
This work is supported by the National Natural Science Foundation of China (No.62441237, No.62172435).
\end{acks}

\bibliographystyle{ACM-Reference-Format}
\bibliography{reference}










\end{document}